\documentclass[sigconf,table]{acmart}

\usepackage{amsthm}
\usepackage{algorithm}
\usepackage{multirow}
\usepackage{algpseudocode}
\usepackage{amsmath}
\usepackage{makecell}

\AtBeginDocument{%
  }

\setcopyright{acmlicensed}
\copyrightyear{2018}
\acmYear{2018}
\acmDOI{XXXXXXX.XXXXXXX}

\acmConference[Conference acronym 'XX]{Make sure to enter the correct
  conference title from your rights confirmation email}{June 03--05,
  2018}{Woodstock, NY}
\acmISBN{978-1-4503-XXXX-X/18/06}

\begin{document}

\title{TimeSoccer: An End-to-End Multimodal Large Language Model for Soccer Commentary Generation}

\author{Ling You}
\authornote{Equal contribution.}
\affiliation{
    \country{}
  \institution{East China Normal University,Shanghai,China}
}
\email{51265901023@stu.ecnu.edu.cn}
\orcid{0009-0000-2422-5106}

\author{Wenxuan Huang}
\affiliation{ \country{}
  \institution{East China Normal University,Shanghai,China}
}
\email{osilly0616@gmail.com}
\orcid{0009-0001-9656-813X}
\authornotemark[1]

\author{Xinni Xie}
\affiliation{  \country{}
  \institution{East China Normal University,Shanghai,China}
}
\orcid{0009-0007-5815-1062}

\author{Xiangyi Wei}
\affiliation{  \country{}
  \institution{East China Normal University,Shanghai,China}
}
\orcid{0009-0006-7630-5558}

\author{Bangyan Li}
\affiliation{  \country{}
  \institution{East China Normal University,Shanghai,China}
}
\orcid{0009-0002-2746-5150}

\author{Shaohui Lin}
\affiliation{  \country{}
  \institution{East China Normal University,Shanghai,China}
}
\email{shlin@cs.ecnu.edu.cn}
\orcid{0000-0003-0284-9940}
\authornotemark[2]

\author{Yang Li}
\affiliation{  
  \institution{East China Normal University,Shanghai,China}
  \country{}
}
\email{yli@cs.ecnu.edu.cn}
\orcid{0000-0001-9427-7665}
\authornote{Corresponding authors}

\author{Changbo Wang}
\affiliation{  \country{}
  \institution{East China Normal University,Shanghai,China}
}
\email{cbwang@cs.ecnu.edu.cn}
\orcid{0000-0001-8940-6418}

\renewcommand{\shortauthors}{You et al.}

\begin{abstract}
%
{
Soccer is a globally popular sporting event, typically characterized by long matches and distinctive highlight moments.
}
{
Recent advances in Multimodal Large Language Models (MLLMs) offer promising capabilities in temporal grounding and video understanding, soccer commentary generation often requires precise temporal localization and semantically rich descriptions over long-form video.
However, existing soccer MLLMs often rely on the temporal a priori for caption generation, so they cannot process the soccer video end-to-end. While traditional approaches follow a two-step paradigm that is complex and fails to capture the global context to achieve suboptimal performance.
}
To solve the above issues, we present \textbf{TimeSoccer}, the first end-to-end soccer MLLM for Single-anchor Dense Video Captioning (SDVC) in full-match soccer videos.
{
\textbf{TimeSoccer} jointly predicts timestamps and generates captions in a single pass, enabling global context modeling across \textbf{45-minute} matches.
To support long video understanding of soccer matches, we introduce \textbf{MoFA-Select}, a training-free, motion-aware frame compression module that adaptively selects representative frames via a coarse-to-fine strategy, and incorporates complementary training paradigms to strengthen the model's ability to handle long temporal sequences.
}{Extensive experiments demonstrate that our \textbf{TimeSoccer} achieves State-of-The-Art (SoTA) performance on the SDVC task in an end-to-end form, generating high-quality commentary with accurate temporal alignment and strong semantic relevance.
}
\end{abstract}




\begin{teaserfigure}
  \includegraphics[width=\textwidth]{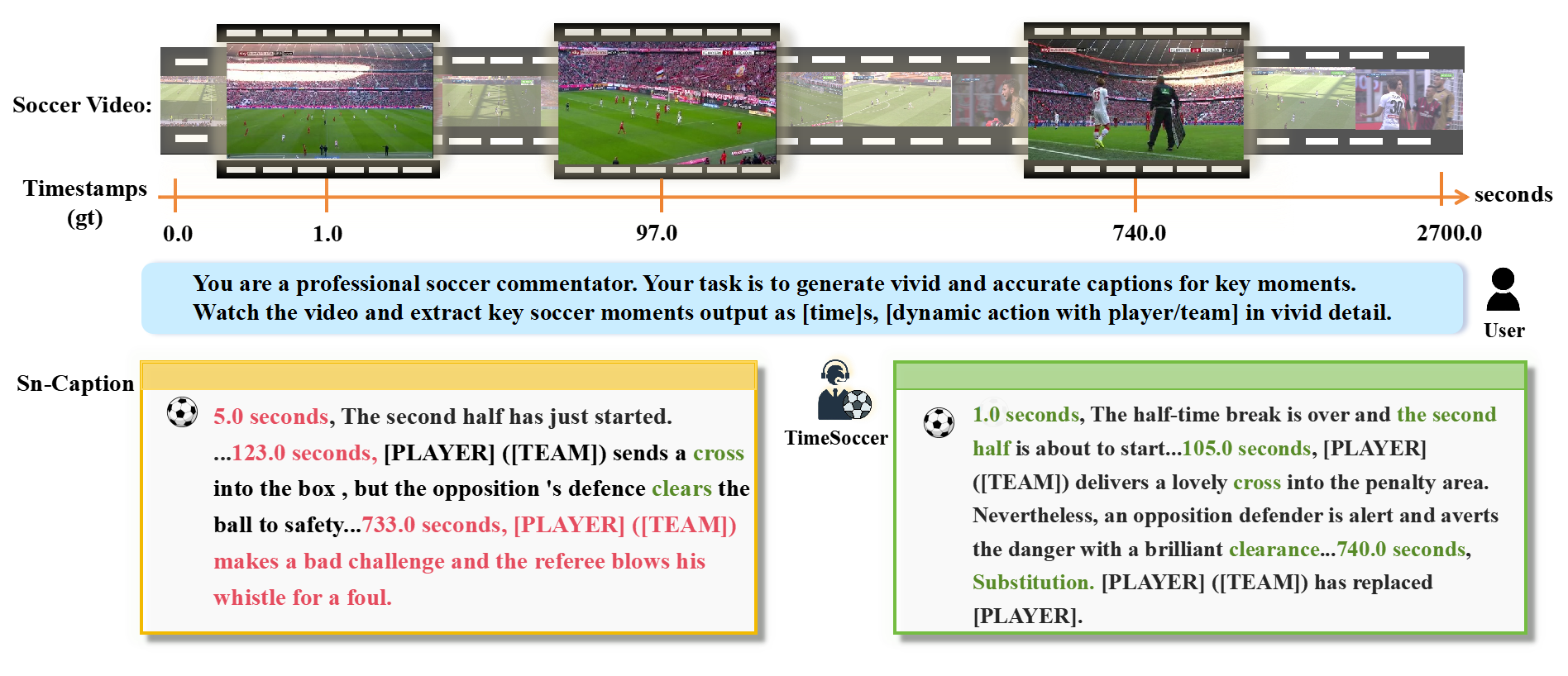}
  \caption{Comparison of soccer commentary performance between \textbf{SoccerNet-Caption}~\cite{mkhallati2023soccernet} and our proposed \textbf{TimeSoccer}. Traditional methods (left) show limited accuracy in temporal alignment and caption quality, while \textbf{TimeSoccer} (right) produces context-aware descriptions with better alignment to ground-truth events. The green (red) text indicates correct (incorrect) predictions in timestamp or content.
}
  \label{fig:teaser}
\end{teaserfigure}



\maketitle

\section{Introduction}
\label{sec:intro}


\begin{figure*}[t]
  \centering
  \includegraphics[width=0.9\linewidth]{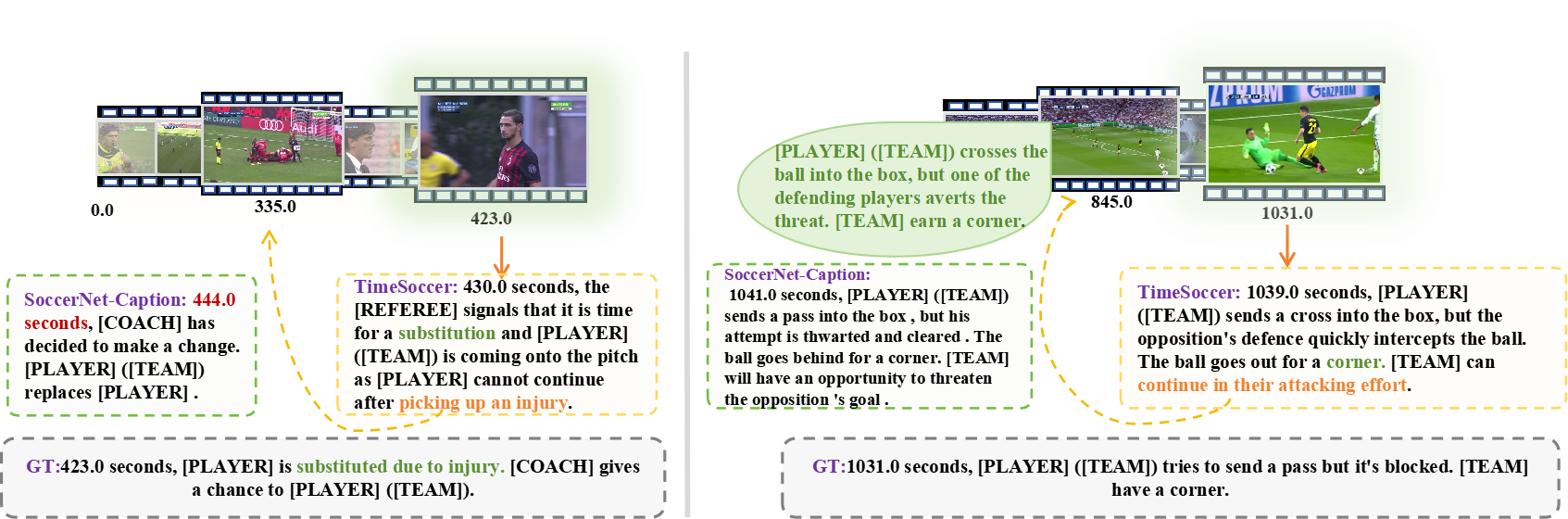}
  \caption{
  An illustration highlighting the global contextual understanding of \textbf{TimeSoccer} compared to SN-Caption~\cite{mkhallati2023soccernet}. 
  \textbf{Left panel}: \textbf{TimeSoccer} correctly attributes a player substitution to a prior injury event; \textbf{Right panel}: \textbf{TimeSoccer} identifies a prior corner kick that occurred earlier in the match and appropriately continues the commentary with “[TEAM] can continue in their attacking effort.” The green (red) text indicates accurate (inaccurate) model-generated timestamps or descriptions.
    }
  \label{fig:global}
\end{figure*}

Soccer is a globally influential sport and remains one of the most popular worldwide.
{With the advancement of soccer-related research~\cite{giancola2018soccernet,cioppa2024soccernet2024challengesresults,cioppa2020multimodal,gao2023trajectory}, increasing efforts have been devoted to soccer video understanding and commentary generation.
}{Recently, Multimodal Large Language Models (MLLMs) have shown remarkable capability in vision tasks, attracting growing interest across various domains~\cite{li2023llava,abu2024using,jin2024chat,weng2024longvlm,song2024moviechat, huang2024dynamic, huang2025vision, li2025llava}.
In particular, MLLMs have made significant achievements in video understanding field~\cite{song2024moviechat,weng2024longvlm, ren2024timechat,wang2025timezero,zeng2024timesuite}, which have opened up new possibilities for soccer commentary generation.
However, effective commentary in soccer matches not only conveys facts but also captures emotional intensity and aligns precisely with key moments, making it a compelling challenge for vision-language research~\cite{mkhallati2023soccernet,rao2024matchtime,rao2024towards}.
}

{In the soccer commentary generation field, recent works have also explored the solution of MLLMs~\cite{cioppa2024soccernet2024challengesresults,rao2024matchtime,rao2024towards}.
Unfortunately, these works focus on optimizing commentary content based on ground-truth temporal segments, without modeling temporal localization, \textit{i.e.}, rely on the temporal a priori for caption generation, making them difficult to apply in real-world soccer commentary scenarios.
}{Some traditional approaches, such as SoccerNet-Caption~\cite{mkhallati2023soccernet}, address soccer commentary generation without relying on predefined timestamps. This setting, known as Single-anchor Dense Video Captioning (SDVC), follows a two-step pipeline: first localizing temporal segments, and then generating the corresponding captions.
}This decoupled design makes commentary generation tightly dependent on the timestamping module, resulting in a pipeline that is not only restrictive but also suboptimal, as each commentary is generated based solely on a short clip, limiting the model's ability to capture global context, as illustrated in Fig.~\ref{fig:global}.

To address the above issue, we propose \textbf{TimeSoccer}, an end-to-end model for the SDVC task, capable of capturing global context and jointly predicting timestamps and commentaries in an end-to-end single pass form for soccer commentary generation. Fig.~\ref{fig:teaser} illustrates a comparison of commentary capabilities between our method and the traditional approach.
{To address the long video understanding problem in soccer matches, we first extend the time-aware large language model TimeChat~\cite{ren2024timechat} by introducing \textbf{MoFA-Select}, a \textbf{Mo}tion-aware, \textbf{F}ine-to-coarse \textbf{A}daptive \textbf{F}rame \textbf{Select}ion module that clusters similar frames and iteratively merges redundancies into a fixed-length representation, aiming to retain semantically important and motion-sensitive frames while reducing inference resource cost and ensuring compatibility with the input constraints of downstream network modules.
}Furthermore, we incorporate two complementary strategies to support long-range temporal modeling: (i) a progressive learning schedule that gradually increases video length during training, helping the model adapt to longer temporal contexts; and (ii) a position embedding extrapolation strategy that enables the model to generalize beyond its original temporal scope. By jointly leveraging these two strategies, the model is better equipped to capture long-range temporal dependencies and maintain semantic coherence throughout the \textbf{full 45-minute} soccer match.
Extensive experiments demonstrate that our approach achieves SoTA performance on the SDVC task, exhibiting high temporal precision and generating semantically high-quality captions. 

Our main contributions are as follows:
\begin{itemize}
\item We propos the first end-to-end soccer MLLM for commentary generation over full-length soccer videos, introducing a new paradigm for the SDVC task with enhanced global context modeling. 
\item We propose \textbf{MoFA-Select}, a training-free frame compression module that selects representative frames via feature similarity through a coarse-to-fine strategy, preserving key information to facilitate long video understanding, and further introduce a progressive training strategy from both learning and architectural perspectives to maintain the performance for the full 45-minute soccer match.
\item Extensive experiments and ablation studies demonstrate that our model achieves SoTA performance on the SDVC task by capturing global context for precise localization and semantically rich commentary.
\end{itemize}

\section{Related Works}

\subsection{MLLMs for Video Understanding}

Recent advances in vision-language modeling have led to the development of powerful architectures~\cite{CLIP, zhai2023SigLIP, li2022blip, li2023blip2, alayrac2022flamingo}, which have achieved impressive results across a wide range of tasks, including classification, image captioning and image-text retrieval. 
Building upon these solid foundations, the field of video understanding has witnessed rapid progress in recent years, with MLLMs achieving remarkable performance on tasks such as multimodal dialogue~\cite{li2023videochat2,fan2024videoagent}, event forecasting~\cite{li2024mm}, time grounding~\cite{ren2024timechat,wang2025timezero,zeng2024timesuite} and dense video captioning~\cite{zhou2024streaming, yang2023vid2seq,ren2024timechat}. In the field of dense video captioning, Video-LLaMA~\cite{zhang2023video} introduces techniques to align visual representations with language prompts, leveraging LLaMA~\cite{touvron2023llama, touvron2023llama2} for generating video descriptions. 
TimeChat~\cite{ren2024timechat} aligns video frames with their corresponding timestamps by injecting temporal information into the image Q-Former~\cite{li2023blip2}, enabling more accurate prediction of temporal segments. 
However, most existing efforts in generating captions from videos focus on open-world scenarios~\cite{song2015tvsum,krishna2017dense,li2023videochat} and overlook the unique challenges of soccer-specific commentary generation, which requires not only precise localization of event occurrences but also the ability to produce accurate and realistic soccer-specific descriptions.

Long video understanding remains a challenging task in computer vision due to the difficulty of maintaining temporal coherence and capturing key information over thousands of frames. 
To address this, several works~\cite{cheng2022xmem,song2024moviechat} have proposed memory-based mechanisms to preserve essential content across time.
For example, XMem~\cite{cheng2022xmem} uses a multi-slot memory architecture with interconnected feature stores to process long videos efficiently, while MovieChat~\cite{song2024moviechat} adopts a hybrid memory design combining short and long-term memory for temporal reasoning. 
In parallel, many token reduction strategies~\cite{bolya2023token,yao2024deco,jiang2025kind} offer insights for long video modeling: ToMe~\cite{bolya2023token} merges similar tokens into shared attention clusters, and G-Prune~\cite{jiang2025kind} prunes redundant tokens via graph-based semantic reasoning. 
These approaches improve the efficiency and scalability of MLLMs for long-form video. 
However, they often overlook motion patterns or rely on complex memory designs, limiting their applicability to real-world sports. 
To this end, we introduce \textbf{MoFA-Select}, a frame-level selection strategy that adaptively retains motion-aware and semantically important frames for long soccer video understanding.

\subsection{Soccer Video Captioning}
Most existing research on soccer video understanding has been conducted on the SoccerNet~\cite{giancola2018soccernet,deliege2021soccernet} dataset family, covering various tasks such as action spotting~\cite{giancola2018soccernet}, player tracking~\cite{cioppa2022soccernet-track}, and commentary generation~\cite{mkhallati2023soccernet, goal, cioppa2024soccernet2024challengesresults}. With the emergence and rapid development of MLLMs, recent works~\cite{rao2024matchtime,rao2024towards} have begun to explore applying MLLMs to soccer commentary generation. 
These methods typically pass short video segments (around 30 seconds) into MLLMs to generate one corresponding caption per clip.
However, they generally overlook the temporal localization aspect, relying on ground-truth timestamps as input.
Even in models that incorporate temporal modeling~\cite{mkhallati2023soccernet}, the dominant paradigm remains a two-stage pipeline: a spotting module is first employed to identify timestamps corresponding to key events, followed by a captioning module that generates captions based on the trimmed video segments. This decoupled design causes commentary generation to be tightly dependent on the timestamping module and prevents the model from capturing the global context.
In contrast, our proposed framework supports the end-to-end prediction of both timestamps and commentary in a single pass, enabling more coherent global understanding and better alignment with the demands of real-world soccer commentary generation.


\begin{figure*}[t]
  \centering
  \includegraphics[width=\linewidth]{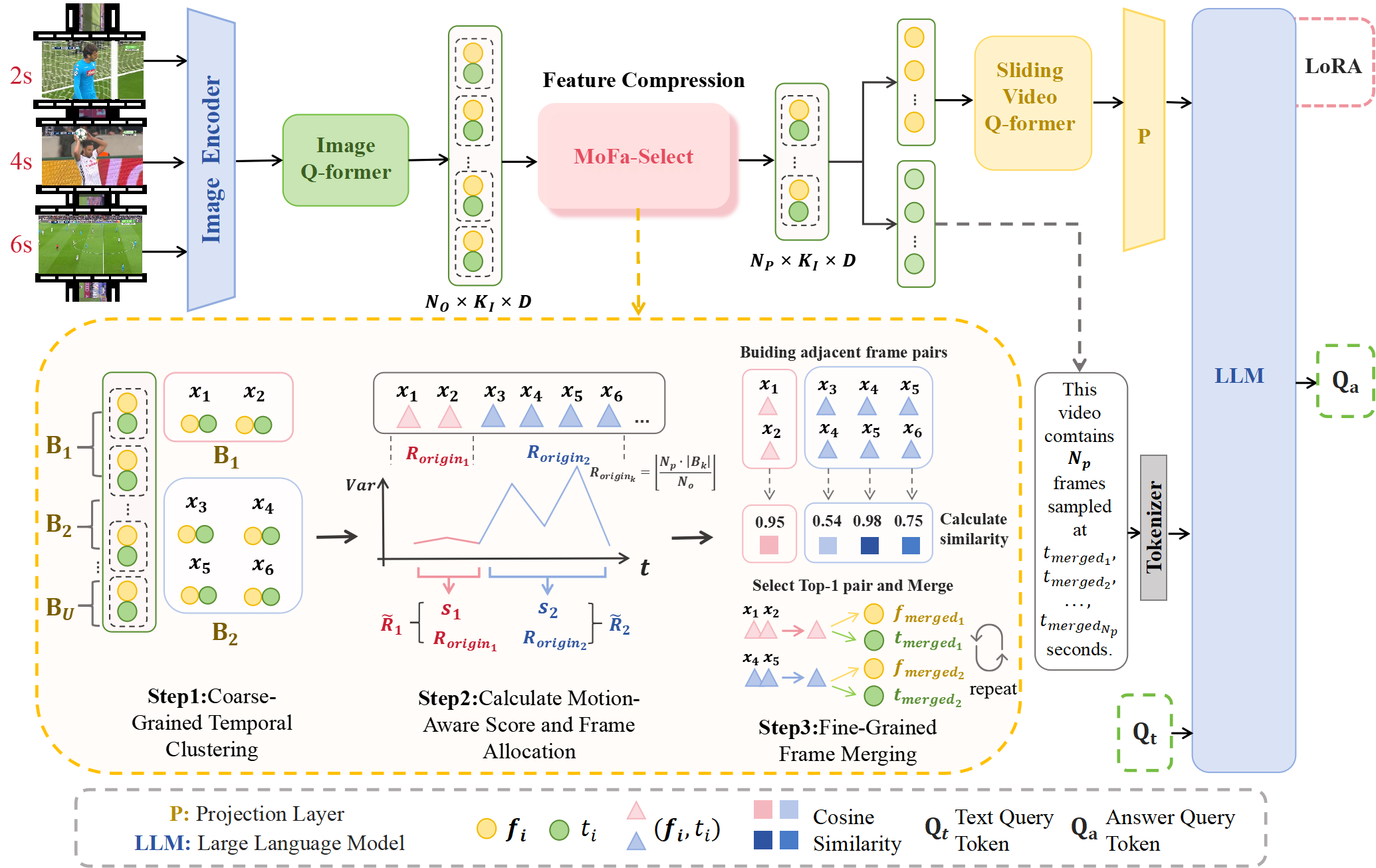}
  \caption{
  Overview of \textbf{TimeSoccer}. Given a full 45-minute soccer video, frame features are extracted using an Image Encoder and Image Q-Former, while timestamps are obtained from the original frame sequence. Both features and timestamps are then processed by the \textbf{MoFA-Select} module, which (a) applies time-constrained K-Means clustering to segment frames, (b) computes motion-aware scores to allocate frame budgets \( \tilde{R}_k \), and (c) merges redundant frames within each segment. The compressed features are passed through a sliding Video Q-Former to generate video tokens, which are concatenated with timestamp-based text tokens and the user query token before being input into the LLM for final prediction.
    }
  \label{fig:pipeline}
\end{figure*}

\section{Method}
\label{sec:method}


In this section, we introduce our proposed model, \textbf{TimeSoccer}, an end-to-end MLLM designed for soccer commentary generation. 
We begin by formulating the problem in Sec.~\ref{sec:problem_formulation}. Sec.~\ref{sec:architecture} outlines the overall pipeline of \textbf{TimeSoccer} and the baseline models it builds upon. 
A detailed description of the proposed \textbf{MoFA-Select} module is provided in Sec.~\ref{sec:mofa}.
Finally, in Sec.~\ref{sec:training}, we introduce our training strategy to further enhance the model's long video understanding capability.





\subsection{Problem Formulation}
\label{sec:problem_formulation}
Given a soccer game video, {\em i.e.}, $\mathbf{V} \in \mathbb{R}^{R \times 3 \times H \times W}$, we aim to build an end-to-end model $\Phi_{\mathrm{TimeSoccer}}$ that identifies all instants \( \hat{\mathbf{T}} \) within the video where a commentary should be anchored and generates a corresponding soccer commentary $\mathbf{C}$ for each detected instant. Each commentary should describe and explain the event occurring around that specific moment in the video, which can be formulated as:
\begin{equation}
    (\hat{\mathbf{T}}, \mathbf{C}) = \Phi_{\mathrm{TimeSoccer}}(\mathbf{V})
\end{equation}
where $(\hat{\mathbf{T}}, \mathbf{C}) = \{(\hat{t_1}, C_1), (\hat{t_2}, C_2), \dots, (\hat{t_n}, C_n)\}$ denotes the set of predicted timestamps and their corresponding commentary captions.

\subsection{Architecture}
\label{sec:architecture}

Our proposed \textbf{TimeSoccer} incorporates several key components, including the time-aware frame visual feature extractor, the similarity-aware frame compression module (\textbf{MoFa-Select}), the sliding video Q-former, and the Large Language model (LLM), as illustrated in Fig.~\ref{fig:pipeline}. Given an input video, the time-aware frame encoder first extracts temporal features from video frames. These features are then processed by the \textbf{MoFA-Select} module, which compares frame-level representations and retains only the most representative keyframe features. The selected keyframe features are subsequently passed through a sliding video Q-former to generate temporally linked video tokens. Finally, these video tokens are fed into the LLM to produce the commentary output.

\noindent \textbf{Baseline.}
We adopt TimeChat~\cite{ren2024timechat} as our baseline due to its strong performance in modeling temporal understanding for long videos. For a given input video $\mathbf{V}$, the model first employs a time-aware frame encoder to extract temporal features from each frame,{\em i.e.}, $\mathbf{F} \in \mathbb{R}^{ R\times K_{I} \times D}$. Specifically, this encoder appends the frame's timestamp—e.g., ``\texttt{This frame is sampled at 2s.}''—as a condition to the Q-Former, enabling it to fuse both visual and temporal information.

To capture temporal relationships across frames, TimeChat utilizes a sliding video Q-former. It introduces a sliding window of length $M_w$, which moves sequentially across the sampled frame features with a stride of $S$. The features within each window are passed into the video Q-former, generating a sequence of video tokens with shape $(R/S) \times k_v \times D$. This mechanism mitigates the issue of excessive compression in long video understanding.

The resulting video tokens $\mathbf{Q}_v$ are projected via a video projection layer and concatenated with the corresponding text tokens $\mathbf{Q}_{t}$ before being fed into a large language model to produce the final output $\mathbf{Q}_{a}$. During training, given an answer sequence $Q_a$ of length $M_a$, the corresponding loss $\mathcal{L}$ is computed as:
\begin{equation}
\mathcal{L} = - \sum_{i=1}^{\mathrm{M_a}} \log P(\mathbf{Q}_a^{(i)} \mid \mathbf{Q}_a^{(<i)}, \mathbf{Q}_v, \mathbf{Q}_t)
\end{equation}

\subsection{MoFA-Select}
\label{sec:mofa}
Although TimeChat demonstrates strong temporal grounding capabilities for long videos, it still struggles to process excessively long inputs (e.g., 45-minute matches) in a single pass. To effectively handle such long-form video inputs while preserving essential motion information, we propose \textbf{MoFA-Select}, a motion-aware, fine-to-coarse adaptive frame selection method.  
This method aims to extract semantically important and motion-sensitive frames from a large pool of sampled frames, while simultaneously reducing memory footprint and ensuring compatibility with the input length constraints of the original sliding video Q-former.

\noindent \textbf{Visual Feature Extraction.}
Given an input video $\mathbf{V} \in \mathbb{R}^{R \times 3 \times H \times W}$, we first sample $N_{o}$ frames and extract their visual features using the time-aware frame feature extractor $\Phi_{\mathrm{extractor}}$. This process can be formulated as:
\begin{equation}
    \mathcal{X}_{N_{o}} = \left\{ \left( \mathbf{f}_{i}, {t}_{i} \right) \;\middle|\; \mathbf{f}_{i} = \Phi_{\mathrm{extractor}}(\mathbf{v}_{i}),\; i = 1, \ldots, N_{o} \right\}
\end{equation}
where $\mathcal{X}_{N_{o}}$ denotes the set of extracted features $\mathbf{f}_{i}$ along with their corresponding timestamps $t_{i}$. Our goal is to compress the initial $N_{o}$ frames into a shorter sequence with a fixed length $N_{p}$.

\noindent \textbf{Coarse-Grained Temporal Clustering.}
Subsequently, we apply a coarse-to-fine compression strategy to capture key information from both global and local perspectives. First, we perform coarse-grained segmentation on the obtained set $\mathcal{X}_{N_{o}}$.
To identify semantically and visually similar temporal segments, we calculate the cosine similarity between each pair of features within the set $\mathcal{X}_{N_{o}}$ as follows:

\begin{equation}
\text{sim}(\mathbf{f}_i, \mathbf{f}_j) = \frac{\mathbf{f}_i \cdot \mathbf{f}_j}{\|\mathbf{f}_i\|_2 \, \|\mathbf{f}_j\|_2}, \quad \forall i, j \in \{1, \ldots, N_o\}
\end{equation}
where \( \text{sim}(\cdot) \) computes the cosine similarity between two feature vectors.
Specifically, we implement a continuous variant of K-Means clustering, which imposes temporal continuity constraints. The frames are initially clustered into $U$ segments by minimizing cumulative cosine distance:

\begin{equation}
\mathcal{L}_{\text{clust}} = \sum_{k=1}^{U} \sum_{t \in \text{cluster}_k} \left(1 - \text{sim}(\mathbf{f}_t, \mathbf{c}_k)\right)
\end{equation}
where $\mathbf{c}_k$ represents the centroid of cluster $k$, and $\mathcal{L}_{\text{clust}}$ denotes the total clustering loss. 
As a result, the set \( \mathcal{X}_{N_{o}} \) is partitioned into subsets \( B_k \) for \( k = 1, \dots, U \).

\noindent \textbf{Motion-Aware Adaptive Frame Allocation.}
To prevent important dynamic events from being mistakenly merged due to high inter-frame similarity, we assign a motion-aware importance weight to each cluster \( B_k \) to ensure that segments with significant motion are appropriately preserved. Specifically, we introduce a motion-aware score \( s_k \) for each temporal cluster \( B_k \), calculated based on the variance within each cluster:

\begin{equation}
s_k = \frac{\mathrm{Var}(\mathbf{f}_t \mid t \in B_k)}{\max_j \mathrm{Var}(\mathbf{f}_t \mid t \in B_j)}.
\end{equation}

After obtaining the motion-aware score $s_k$ for each cluster \( B_k \), we proportionally allocate frames to each segment based on its relative motion contribution.
The number of frames allocated to each cluster \( B_k \), denoted as \( R_k \), is determined as follows:

\begin{equation}
\label{allocate}
R_k = \max\left(1, \min\left(\left\lfloor \frac{N_{P} \cdot |B_k|}{N_{o}} \right\rfloor + s_k \cdot \left\lfloor \frac{N_{P} \cdot |B_k|}{N_{o}} \right\rfloor, |B_k| \right)\right)
\end{equation}
where \( R_{\text{origin}} = \left\lfloor \frac{N_{P} \cdot |B_k|}{N_{o}} \right\rfloor \) denotes the initial number of frames that would be allocated to cluster \( B_k \) based on its relative size before applying motion-aware scaling. Based on Equation~\ref{allocate}, once the target number of frames \( R_k \) for each cluster is determined, we proportionally scale them to ensure that the total number of assigned frames equals the final compressed length \( N_p \) and denote the scaled targets as \( \tilde{R}_k \). Such scaling guarantees a balanced and motion-aware allocation, prioritizing dynamic segments while strictly matching the desired compressed length.

\noindent \textbf{Fine-Grained Frame Merging.}
Within each cluster \( B_k \), inspired by Moviechat~\cite{song2024moviechat}, we iteratively merge features in the adjacent frames by selecting pairs of frames with maximum similarity. To preserve motion dynamics, we compute a motion penalty based on the variance between merged frames:

\begin{equation}
\Delta_{\text{motion}}(i) = \mathrm{Var}(\{\mathbf{f}_i, \mathbf{f}_{i+1}\})
\end{equation}
where \( \Delta_{\text{motion}}(i) = \mathrm{Var}(\{\mathbf{f}_i, \mathbf{f}_{i+1}\}) \) denotes the motion penalty function between adjacent frames \( i \) and \( i+1 \), computed as the variance of their feature representations.
If $\Delta_{\text{motion}}$ exceeds a predefined threshold \( \delta \), we discard redundant frames instead of merging them to preserve important motion information. The threshold \( \delta \) is empirically set to 0.3 based on validation performance.
Otherwise, we merge adjacent frames by averaging their feature representations as follows:

\begin{equation}
\mathbf{f}_\text{merged} = \frac{\mathbf{f}_i + \mathbf{f}_{i+1}}{2}, \quad t_\text{merged} = \frac{t_i + t_{i+1}}{2}.
\end{equation}

The merging process is repeated iteratively until each cluster \( B_k \) reaches its allocated number of frames \( \tilde{R}_k \). By concatenating the retained features from all clusters in temporal order, we obtain the final compressed representation \( \mathcal{X}_{N_{p}} \).

\noindent \textbf{Conclusion.}
After applying \textbf{MoFA-Select}, the compressed representation \( \mathcal{X}_{N_p} = \{(\mathbf{f_1}, t_1), \dots, (\mathbf{f_{N_p}}, t_{N_p})\} \) preserves key motion segments while maintaining temporal coherence, resulting in a balanced and informative fixed-length sequence for downstream processing. Specifically, the visual features \( \{\mathbf{f_1}, \dots, \mathbf{f_{N_p}}\} \) are passed through a sliding Video Q-Former to produce video tokens, while the corresponding timestamps \( \{t_1, \dots, t_{N_p}\} \) are formatted into a textual summary of the form: “\texttt{This video contains \( N_p \) frames sampled at \( t_1 \), \( t_2 \), ..., \( t_{N_p} \) seconds.}” These timestamp-based text tokens are then concatenated with the user query and video tokens, and jointly fed into the LLM to generate the final response.

\subsection{Training Process}
\label{sec:training}
To further enable our model to understand a full-length soccer video (typically 45 minutes), we enhance its long video modeling capability from both the learning and architectural perspectives in the training process. \textbf{From the learning perspective}, we adopt a progressive training strategy, where the model is gradually fine-tuned on increasingly longer videos, allowing it to incrementally build the ability to comprehend extended temporal contexts. \textbf{From the architectural perspective}, to retain more temporal details by reducing compression, we employ a position embedding extrapolation strategy based on periodic replication, where the pretrained positional embeddings are cyclically repeated to cover the desired length. This approach ensures seamless compatibility with pretrained models while enabling flexible adaptation to varying input lengths.

\begin{table*}[t]
\caption{
Quantitative comparison of temporal localization and caption quality across different methods on the \textit{Soccernet-Caption} dataset.
The best results are \textbf{bolded} and the second best results are \underline{underlined} in all following tables.
M-Score and C-Score denote Qwen2.5-VL-72B-Instruct evaluations for match consistency and overall commentary quality. “SN-Caption+X” indicates that the model generates commentary based on timestamps predicted by SN-Caption, while “3-minute via 15× inference” refers to generating the final match-level commentary by performing 15 separate inferences on consecutive 3-minute segments.
$\dag$ indicates that SN-Caption requires generating commentary for each predicted temporal segment individually, resulting in multiple forward passes.
}
\centering
\large
\resizebox{1.00\linewidth}{!}{
\begin{tabular}{@{}llcccccccccc@{}}  
\toprule
\multicolumn{2}{c}{} 
& \multicolumn{5}{c}{\textbf{Temporal Metrics $\uparrow$}} 
& \multicolumn{3}{c}{\textbf{Caption Quality $\uparrow$}} 
& \multicolumn{2}{c}{\textbf{LLM Score $\uparrow$}} \\
\cmidrule(lr){3-7} \cmidrule(lr){8-10} \cmidrule(lr){11-12}
\multicolumn{2}{c}{\textbf{Method}} 
& P@0.3 & P@0.5 & P@0.7 & P@0.9 & F1 
& CIDEr & METEOR & SODA\_c 
& M-S & C-S \\
\midrule
\rowcolor{cyan!15}
\multicolumn{12}{c}{\textit{Soccer-specific Baselines}} \\
& $\dag$(CVPR23) Sn-Caption~\cite{mkhallati2023soccernet} & 12.5 & 7.2 & 3.0 & 1.1 & \textbf{8.9} & \textbf{11.0} & 4.9 & \textbf{3.9} & 3.41 & \underline{4.30} \\
& \makecell[l]{(CVPR23) SN-Caption~\cite{mkhallati2023soccernet}+(EMNLP24) MatchTime~\cite{rao2024matchtime}} & 12.5 & 7.2 & 3.0 & 1.1 & \textbf{8.9} & 7.9 & 4.1 & 3.2 & 3.55 & 4.21\\
& \makecell[l]{(CVPR23) SN-Caption~\cite{mkhallati2023soccernet}+(CVPR25) UniSoccer~\cite{rao2024towards}} & 12.5 & 7.2 & 3.0 & 1.1 & \textbf{8.9} & 8.0 & 4.5 & \underline{3.3} & \underline{3.90} & 4.29\\
\rowcolor{cyan!15}
\multicolumn{12}{c}{\textit{General Video MLLMs}} \\
& (EMNLP23) Video-LLaMA~\cite{zhang2023video} & 0.0 & 0.0 & 0.0 & 0.0 & 0.1 & 0.0 & 0.0 & 0.0 & 1.65 & 1.90 \\
& (CVPR24) MovieChat~\cite{song2024moviechat} & 2.5 & 2.2 & 2.0 & 0.1 & 2.5 & 0.0 & 0.1 & 0.1 & 0.84 & 1.53 \\
& (Arxiv24) LLaVA-Video-72B-QWen2~\cite{zhang2024video} & 2.2 & 2.2 & 0.1 & 0.1 & 1.2 & 0.0 & 0.0 & 0.1 & 1.57 & 1.91 \\
& (TMLR25) LLaVA-Onevision-QWen2-72B~\cite{li2024llava} & 3.1 & 2.1 & 0.1 & 0.1 & 1.4 & 0.0 & 0.1 & 0.2 & 3.00 & 3.44 \\
& (CVPR24) TimeChat (ft)~\cite{ren2024timechat} & \underline{13.4} & \underline{9.0} & \underline{4.1} & \underline{1.6} & 8.2 & 5.5 & \underline{5.1} & 2.6 & 4.50 & 5.02\\
\rowcolor{cyan!15}
\multicolumn{12}{c}{\textit{Ours}} \\
& \textbf{TimeSoccer (45-minute)}  & \textbf{17.0} & \textbf{11.0} & \textbf{6.0} & \textbf{3.4} & \underline{8.8} & \underline{8.3} & \textbf{6.2} & 2.7 & \textbf{5.03} & \textbf{5.14} \\
\midrule
& \color[gray]{0.4}{\textbf{TimeSoccer (3-minute via 15$\times$ inference)}} & \color[gray]{0.4}{22.5} & \color[gray]{0.4}{15.4} & \color[gray]{0.4}{8.0} & \color[gray]{0.4}{3.8} & \color[gray]{0.4}{12.0} & \color[gray]{0.4}{13.8} & \color[gray]{0.4}{5.5} & \color[gray]{0.4}{4.3} & \color[gray]{0.4}{4.12} & \color[gray]{0.4}{6.18} \\
\bottomrule
\end{tabular}
}
\label{tab:quantitative}
\end{table*}

\begin{figure*}[t]
  \centering
  \includegraphics[width=0.95\linewidth]{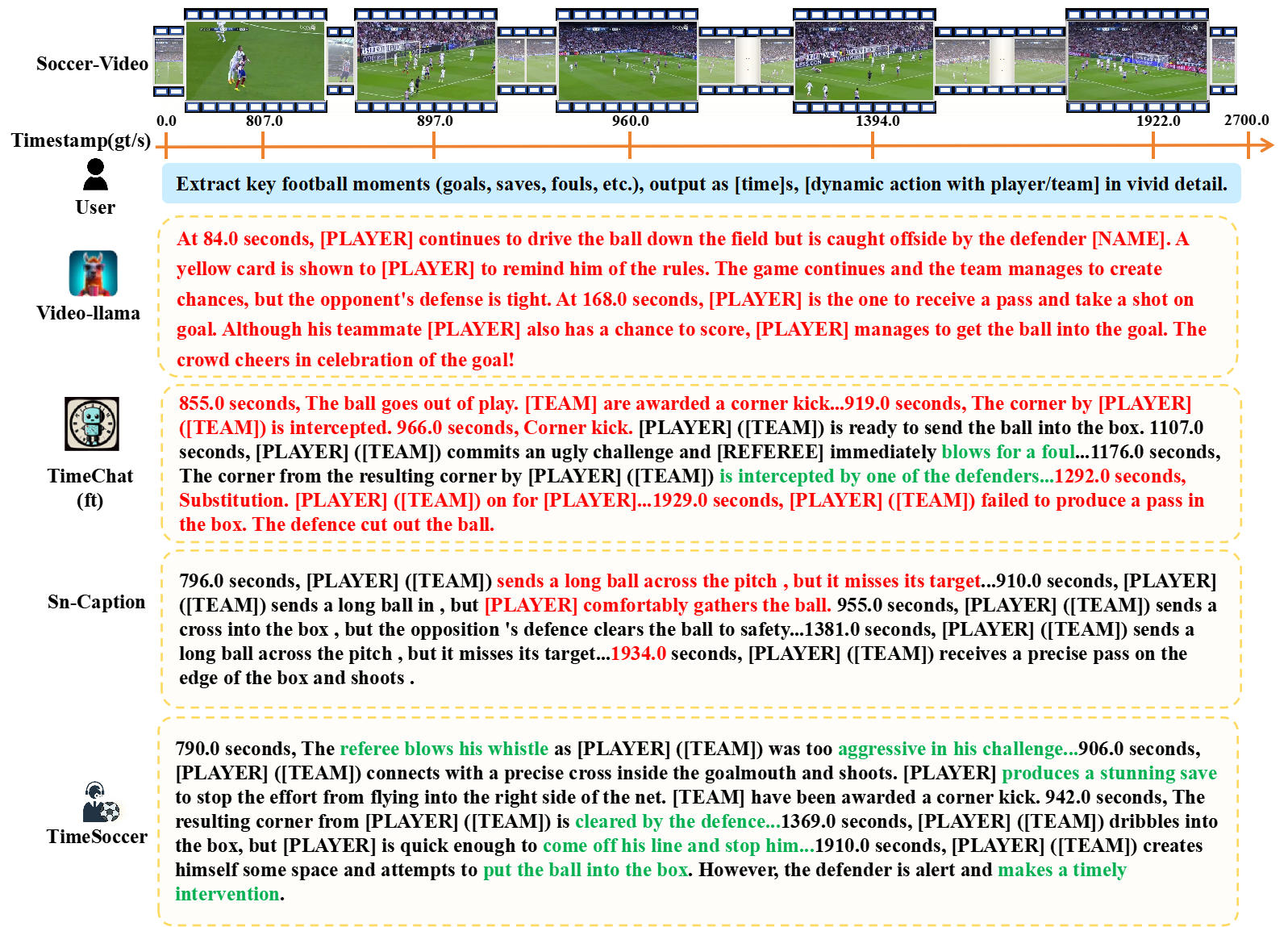}
  \caption{Quality comparison results on different methods. 
  \textbf{TimeSoccer} demonstrates its advantages from multiple perspectives:  
(i) more accurate timestamp alignment;  
(ii) improved event descriptions;  
(iii) richer, more realistic commentary resembling professional broadcasts. 
Black text denotes outputs that are reasonably close to the ground truth in terms of either timing or semantics, while green highlights semantically accurate descriptions, and red marks incorrect or irrelevant content.
  }
  \label{fig:compare}
\end{figure*}

\section{Experiments}
\subsection{Implementation Details}
We adopt ViT-G/14 from EVA-CLIP~\citep{Sun2023EVACLIPIT} as the visual encoder and LLaMA-2 (7B)~\citep{touvron2023llama2} as the language backbone. All remaining modules are initialized from TimeChat~\citep{ren2024timechat}. We fine-tune the Image Q-Former and Video Q-Former separately using the original \textit{Soccernet-Caption} dataset, while keeping both the Vision Transformer (ViT) and the Large Language Model (LLM) frozen, as illustrated in Fig~\ref{fig:pipeline}. To better adapt the LLM to our video understanding task, we employ the parameter-efficient fine-tuning method LoRA, with a rank of 32. The number of clusters in the \textbf{MoFA-Select} module is set to 6. The \( \delta \) is 0.3.
All experiments are conducted on 4 × NVIDIA H20 GPUs (96 GB). 
\subsection{Evaluation Setups}
\noindent \textbf{Dataset.}
We fine-tune and evaluate our model on the \textit{Soccernet-Caption} dataset. Specifically, we divide the dataset into 422 training videos and 49 testing videos, following the train/test split defined in MatchTime~\citep{rao2024matchtime}.
Additionally, we replace the original timestamp annotations with the refined temporal labels provided by MatchTime to ensure higher alignment accuracy.

\noindent \textbf{Evaluation Metrics.}
For temporal evaluation, we extend both ground-truth and predicted timestamps by 5 seconds on each side and compute IoU at thresholds of 0.3, 0.5, 0.7, and 0.9 to measure alignment accuracy. We also report the F1 score to reflect the balance between precision and recall. For caption quality, we adopt CIDEr~\cite{vedantam2015cider}, METEOR~\cite{banerjee2005meteor}, and SODA\_c~\cite{Fujita2020SODASO}.
Previous evaluations of commentary quality have primarily focused on sentence-level alignment between predicted and reference captions, which overlooks the overall coherence and informativeness of the generated commentary. To address this limitation, we employ Qwen2.5-VL-72B-Instruct~\cite{bai2025qwen2} to conduct a more comprehensive assessment through two metrics: match-level semantic alignment (M-S) and comprehensive commentary quality (C-S). Each dimension is rated on a scale from 1 to 10.


\subsection{Quantitative Evaluation}
\label{sec:sota}
To verify the effectiveness of our approach, we compare it with the SoTA SDVC model, SoccerNet-Caption~\cite{mkhallati2023soccernet}, as well as other recent soccer commentary generation methods~\cite{rao2024matchtime,rao2024towards}. Since these methods typically rely on ground-truth timestamps, we feed them with the predicted timestamps from SoccerNet-Caption to ensure a fair comparison. 
We conduct evaluations under both 3-minute and full 45-minute settings. While the 45-minute setting yields slightly lower overall metric scores compared to the 3-minute version, it enables end-to-end inference on the entire match video. In contrast, the 3-minute setup requires 15 separate forward passes and thus cannot be considered a fully end-to-end approach. Therefore, we adopt the 45-minute setting as a representative evaluation protocol and compare it with other baselines to evaluate the holistic performance of our model.
Furthermore, we include comparisons with recent strong end-to-end video understanding models based on MLLMs~\cite{zhang2023video,maaz2024video}, using the same prompt setup as our model, to further demonstrate the superiority of our method in both temporal localization and commentary generation. Quantitative results are presented in Tab.~\ref{tab:quantitative}.

\noindent \textbf{Temporal Localization Accuracy.}
As shown in Tab.~\ref{tab:quantitative}, within soccer-specific baselines, SN-Caption relies solely on a feature extractor, aggregator, and classification head to predict caption probabilities for each frame, which limits its robustness in precise temporal grounding. Building upon SN-Caption, both MatchTime and UniSoccer inherit its timestamp predictions without introducing explicit temporal modeling, thereby constraining their generalizability and limiting their effectiveness in diverse scenarios.
For general-purpose video MLLMs, most methods compress frame features into fixed-length representations, making them struggle to achieve accurate temporal localization.
In contrast, our method outperforms the second-best SN-Caption by +4.5, +3.8, +3.0, and +2.3 on Precision@0.3, 0.5, 0.7, and 0.9 respectively in 45-minute temporal grounding. This highlights our model’s ability to select salient features and accurately localize key moments, significantly enhancing the temporal understanding capability of MLLMs for long videos.


\noindent \textbf{Caption Quality Evaluation.}
As shown in Tab.~\ref{tab:quantitative}, inaccurate temporal localization significantly affects commentary quality, resulting in a low captioning performance for SN-Caption, MatchTime, and UniSoccer. Among them, SN-Caption yields slightly better results, potentially due to the shared parameters between its captioning and spotting modules beyond the classification head, which may offer better temporal compatibility.
For general-purpose video MLLMs, the absence of soccer-specific adaptation and inadequate modeling of long-form temporal structures significantly limit their effectiveness in generating precise and contextually rich soccer captions.
In contrast, our method achieves remarkable performance gains, surpassing the previous SoTA by +2.8 CIDEr, +0.6 METEOR, and +0.4 SODA\_c on 3-minute videos. Furthermore, it remains competitive on full 45-minute videos, demonstrating robust capabilities in long-form temporal understanding.


\noindent \textbf{Model Versatility.}
We evaluate the quality of generated commentary using Qwen2.5-VL-72B-Instruct, focusing on two aspects: match-level semantic alignment (M-S) and comprehensive commentary quality (C-S).
As shown in Tab.~\ref{tab:quantitative}, our method outperforms SN-Caption by +1.62 and +0.84 on M-S and C-S respectively in the 45-minute setting, benefiting from the language model’s strong narrative capability. Moreover, the 45-minute setting yields higher match-level alignment than the 3-minute setting, suggesting that full-length generation helps capture global context more effectively.

\begin{table*}[t]
\caption{
Comparison of \textbf{MoFA-Select} with standard SFT and baseline compression methods on the \textit{Soccernet-Caption} dataset in the 45-minute video setting.
}
\label{tab:mofa_ablation}
\centering
\large
\resizebox{0.75\linewidth}{!}{
\begin{tabular}{@{}llccccc|ccc@{}}
\toprule
\multicolumn{2}{c}{} 
& \multicolumn{5}{c|}{\textbf{Temporal Metrics $\uparrow$}} 
& \multicolumn{3}{c}{\textbf{Caption Quality $\uparrow$}} \\
\cmidrule(lr){3-7} \cmidrule(lr){8-10}
\textbf{} & \textbf{Method} 
& P@0.3 & P@0.5 & P@0.7 & P@0.9 & F1 
& CIDEr & METEOR & SODA\_c \\
\midrule
& TimeChat~\cite{ren2024timechat}              & 0.2 & 0.2 & 0.2 & 0.0 & 0.0 & 0.0 & 0.0 & 0.0 \\
& TimeChat (ft)~\cite{ren2024timechat}                   & 13.4 & 9.0 & 4.1 & 1.6 & 8.2 & 5.5 & 5.1 & 2.6 \\
& Ours (G-Prune~\cite{jiang2025kind})            & 15.5 & 9.9 & 4.7 & 2.4 & 8.5 & 7.7 & 5.8 & 2.7  \\
& Ours w/o (Fine Stage~\cite{song2024moviechat})               & 13.8 & 9.4 & 4.7 & 2.4 & 8.0 & 7.3 & 5.4 & 2.6  \\
& Ours w/o Coarse Stage & 15.0 & 9.9 & 4.6 & 2.2 & 8.5 & 7.5 & 5.6 & 2.8 \\
& Ours w/o Motion-Aware & 15.8 & 10.5 & 5.1 & 2.6 & 8.5 & 6.1 & 2.9 & 4.1  \\
& Ours w/o time-merge & 12.9 & 9.0 & 4.8 & 2.6 & 7.3 & 7.1 & 4.8 & 2.2  \\
& \textbf{Ours (Full MoFA-Select)}         & \textbf{17.0} & \textbf{11.0} & \textbf{6.0} & \textbf{3.4} & \textbf{8.8} & \textbf{8.3} & \textbf{6.2} & \textbf{2.7}  \\
\bottomrule
\end{tabular}
}
\end{table*}
\begin{table*}[t]
\caption{
Ablation Study on Training Paradigms and Positional Encoding Extensions for Full-Match Video Understanding on the \textit{Soccernet-Caption} dataset.
}
\label{tab:ablation_training}
\centering
\large
\resizebox{0.85\linewidth}{!}{
\begin{tabular}{@{}llcccc|ccc@{}}
\toprule
\multicolumn{2}{c}{} 
& \multicolumn{4}{c|}{\textbf{Temporal Metrics $\uparrow$}} 
& \multicolumn{3}{c}{\textbf{Caption Quality $\uparrow$}} \\
\cmidrule(lr){3-6} \cmidrule(lr){7-9}
\textbf{} & \textbf{Method} 
& P@0.3 & P@0.5 & P@0.7 & P@0.9 
& CIDEr & METEOR & SODA\_c \\
\midrule
\rowcolor{cyan!15}
\multicolumn{9}{c}{\textit{Direct Training}} \\
& Direct (45-min Only) & 13.4 & 9.0 & 4.1 & 1.6 & 5.5 & 5.1 & 2.6 \\
& Direct (45-min + Interpolated PosEnc) & 13.2 & 8.9 & 5.0 & 2.6 & 6.7 & 4.5 & 1.2 \\
& Direct (45-min + Repeated PosEnc) & 15.4 & 10.8 & 5.2 & 3.0 & 8.7 & 4.6 & 0.9 \\
\rowcolor{cyan!15}
\multicolumn{9}{c}{\textit{Progressive Training}} \\
& Progressive (3→15→45-min) & 13.8 & 9.0 & 4.7 & 2.7 & 8.6 & 4.6 & 1.2 \\
& Progressive (3→15→45-min + Repeated PosEnc) & 15.3 & 10.3 & 4.9 & 2.2 & 9.2 & 5.7 & 2.6 \\
& \textbf{Progressive + MoFA-Select (Full Model)} & \textbf{17.0} & \textbf{11.0} & \textbf{6.0} & \textbf{3.4} & \textbf{8.3} & \textbf{6.2} & \textbf{2.7} \\
\bottomrule
\end{tabular}
}
\end{table*}

\subsection{Ablation Study}
We conduct a series of ablation experiments to evaluate the effectiveness of key components in our framework. Specifically, we assess the impact of the \textbf{MoFA-Select} module and the complementary training strategies. We further investigate the effect of varying key hyperparameters in the appendix. These studies provide insights into how each part contributes to the overall effectiveness of our method.


\noindent\textbf{Impact of MoFA-Select.}
As shown in Tab.~\ref{tab:mofa_ablation}, when \textbf{MoFA-Select} is not used and the model is fine-tuned solely on the \textit{SN-Caption dataset}, the limited number of sampled frames leads to suboptimal identification of key moments. This results in performance drops under stricter temporal IoU thresholds (0.7 and 0.9), as well as lower CIDEr and METEOR scores (-1.8, -0.6, -2.8, -1.1). 
Both G-Prune~\cite{jiang2025kind} and \textbf{MoFA-Select} enable the model to sample more frames and select the most informative ones, thereby improving the localization of key timestamps and semantic content. 
However, \textbf{MoFA-Select} consistently outperforms G-Prune, as the latter relies solely on global similarity propagation and tends to retain redundant segments.
Furthermore, we conduct a detailed ablation on the four components of MoFA-Select. 
The results show that removing any of them leads to a performance drop, highlighting the necessity of each component in achieving effective long video understanding.

\noindent\textbf{Impact of the progressive training strategy.}
As illustrated in Tab.~\ref{tab:ablation_training}, our progressive training strategy gradually enhances the model’s ability to understand long-form videos, achieving notable improvements over direct training in terms of CIDEr and METEOR, SODA\_c, which demonstrates its effectiveness in facilitating long-range temporal modeling.
Moreover, extending the positional encoding improves performance under both direct and progressive training setups. We further compare two extension approaches: interpolation and periodic replication. 
As shown in the results, periodic extension outperforms interpolation, which may be attributed to the excessive smoothing introduced when interpolating far beyond the original length. This smoothing distorts positional semantics, whereas periodic replication preserves the original embedding distribution, leading to more robust performance.

    

    

\subsection{Qualitative Comparison}
As shown in Fig.~\ref{fig:compare}, we compare the performance of various models on the SDVC task, including VideoLLaMA, TimeChat fine-tuned on 45-minute videos, SN-Caption and our proposed \textbf{TimeSoccer}. The results highlight the following advantages of \textbf{TimeSoccer}:  
(i) \textbf{Improved temporal grounding}: Equipped with the \textbf{MoFA-Select} module, \textbf{TimeSoccer} effectively captures key visual and temporal cues, enabling more accurate alignment between the generated commentary and key events. 
(ii) \textbf{Enhanced contextual coherence}: The enhanced temporal grounding of \textbf{TimeSoccer} allows it to better align commentary with event progression, resulting in more accurate and logically consistent descriptions.
(iii) \textbf{Richer semantic content}: The remarkable ability of MLLM helps \textbf{TimeSoccer} produce more detailed and context-aware descriptions, improving both readability and domain relevance. 
More analysis results can be found in the appendix.

\section{Conclusion}
In this paper, we present \textbf{TimeSoccer}, the first end-to-end framework for Single-anchor Dense Video Captioning (SDVC) in full-length soccer matches. 
Unlike previous methods that rely on ground-truth timestamps or adopt a two-stage paradigm based on short video clips, \textbf{TimeSoccer} jointly predicts temporal segments and generates captions in a single forward pass. This end-to-end design enables holistic modeling of long-range temporal context and allows direct inference over full 45-minute soccer matches.
To support efficient long video understanding, 
We propose \textbf{MoFA-Select}, a training-free, motion-aware frame compression module that adaptively selects representative frames using a coarse-to-fine strategy.  
Finally, we adopt a progressive training strategy to further strengthen the model’s temporal reasoning capabilities.  
Extensive experiments demonstrate that \textbf{TimeSoccer} outperforms existing baselines in both temporal localization and commentary quality.


\bibliographystyle{ACM-Reference-Format}
\bibliography{arxiv}

\end{document}